\documentclass[journal]{IEEEtran}

\usepackage{amsmath,amssymb,amsfonts}
\usepackage{algorithmic}
\usepackage{algorithm}
\usepackage{graphicx}
\usepackage{textcomp}
\usepackage{cite}
\usepackage{url}
\usepackage{xcolor}
\usepackage{verbatim}
\usepackage{booktabs}
\usepackage{tabularx}
\usepackage{array}
\usepackage{placeins}
\usepackage{float}
\usepackage{makecell}
\usepackage{multirow}
\usepackage[figuresright]{rotating}
\usepackage[caption=false,font=footnotesize]{subfig}
\usepackage{stfloats}
\usepackage{siunitx}
\usepackage[colorlinks=true, linkcolor=blue, citecolor=blue, urlcolor=blue]{hyperref}
\begin{document}

\title{SpecLST: Spectral-Enhanced Long-Short Term Spatio-Temporal Multimodal Modeling for Traffic Forecasting}


\author{Kenan Kang, Matthew M.Y. Kuo, and Weihua Li%
\thanks{Kenan Kang and Matthew M.Y. Kuo are with the Department of Computer Science and Software Engineering, Auckland University of Technology, Auckland 1010, New Zealand (e-mail: kenan.kang@autuni.ac.nz; matthew.kuo@aut.ac.nz).}%
\thanks{Weihua Li is with the Department of Data Science and Artificial Intelligence, Auckland University of Technology, Auckland 1010, New Zealand (e-mail: weihua.li@aut.ac.nz).}%
\thanks{Corresponding author: Kenan Kang.}}

\markboth{Preprint}%
{Kang \MakeLowercase{\textit{et al.}}: SpecLST for Traffic Forecasting}

\maketitle

\begin{abstract}
Modeling multi-scale dynamics in urban road networks is essential for intelligent transportation systems but is often hindered by the computational cost of long-range historical context and the challenge of fusing heterogeneous modalities. Existing spatio-temporal frameworks face a trade-off between sequence length and memory overhead, while frequently neglecting the transient effects of exogenous factors such as road incidents and weather. To address these limitations, we propose SpecLST, a frequency-enhanced long-short-term spatio-temporal multimodal framework. SpecLST leverages a fast Fourier transform-based frequency-domain representation module to efficiently encode multi-week historical dependencies, bypassing the memory bottlenecks of standard attention mechanisms over raw long sequences. Heterogeneous context is integrated through attention-based fusion, followed by a sparse Mixture-of-Experts backbone with graph augmentation for spatio-temporal representation learning. Additionally, a frequency-enhanced residual head isolates high-frequency transients, improving sensitivity to abrupt traffic anomalies. At the 60-minute horizon, SpecLST achieves mean relative error reductions of 14.10\% in MAE, 10.35\% in RMSE, and 15.78\% in MAPE across eight neural forecasting baselines and four datasets, with the dataset- and metric-specific average reductions ranging from 5.63\% to 22.21\%. 
\end{abstract}


\section{Introduction}
\label{sec:introduction}

Accurate traffic forecasting underpins modern Intelligent Transportation Systems (ITS), supporting route guidance, congestion mitigation, and vehicle dispatching \cite{VLAHOGIANNI20143,Yuan2021}. It also drives operational tasks such as adaptive signal control, demand-responsive transit, and incident response. Multimodal data is now widely accessible, including PeMS loop detector readings, social media incident reports, and high-resolution weather feeds. As urbanization accelerates, traffic flow remains highly non-linear and non-stationary, which makes precise multi-horizon prediction difficult in practice. Different tasks demand different forecasting capabilities: signal control benefits from multi-week periodicity, incident response from fusing exogenous signals with traffic, and short-term replanning from rapid response to abrupt deviations. Current models rarely cover more than one of these at once.

Over the past decade, the field has shifted from statistical methods such as VAR \cite{VAR_baselines} to deep learning. Spatio-Temporal Graph Neural Networks (STGNNs), such as DCRNN \cite{li2018dcrnn} and MTGNN \cite{wu2020mtgnn}, learn the non-Euclidean connectivity of road networks. More recent Transformer models, such as STAEformer \cite{staeformer} and LSTTN \cite{lsttn}, extend the temporal receptive field through self-attention. Three bottlenecks remain. First, existing models face a ``Context-Complexity Dilemma'': long-term history (e.g., several weeks) carries useful periodic priors, but the quadratic cost of self-attention restricts most models to roughly one hour of input. Second, although recent multimodal frameworks incorporate weather and calendar features \cite{9363197}, other exogenous signals such as social-media incidents and short-term weather volatility remain underused, and are rarely combined with long-history representations in a single architecture. Third, most backbones treat the short-term input as a smooth signal and respond slowly to abrupt, incident-driven deviations. The high-frequency content of recent observations, which can provide useful signatures of transient anomalies, is rarely modelled explicitly.

To bridge these gaps, we propose SpecLST, a novel spectral-enhanced spatio-temporal Mixture-of-Experts framework. The core philosophy of our approach is to decouple traffic patterns into stable periodic backbones and volatile transient residuals through a dual-branch spectral analysis. Concretely, SpecLST first maps two weeks of historical traffic into the frequency domain and represents it as a compact set of low-frequency coefficients, aligned with the recent short-term window through phase rotation. Because urban traffic exhibits strong daily and weekly periodicity, the Fourier basis is a natural fit: a small number of low-frequency coefficients can retain the dominant periodic information in the multi-week signal, allowing the model to avoid directly applying self-attention to raw long sequences while preserving the periodic structure that matters for forecasting. The resulting periodic embedding is then combined with short-term traffic, weather, calendar, and, where available, curated incident-event modalities through a dedicated attention-based context-fusion module before entering the spatio-temporal backbone. The backbone consists of stacked blocks that alternate feed-forward and sparse Mixture-of-Experts layers, with spatial attention operating on a fused adjacency that augments the static road topology with a learnable sparse component. Finally, a lightweight frequency-enhanced residual head (FRH) consumes the high-frequency power spectrum of a separate recent window and produces a per-horizon, gated residual correction on top of the backbone's prediction, sharpening responsiveness to transient anomalies. 

The main contributions of this paper are as follows:
\begin{itemize}
    \item We propose SpecLST, to our knowledge the first traffic forecasting framework that jointly integrates a compact frequency-domain long-history representation, attention-based multimodal context fusion, and a per-horizon high-frequency residual head within a  unified architecture.
    \item We extend the public PEMS-BAY22 traffic benchmark~\cite{PEMS-BAY-22} with aligned exogenous context, including weather, calendar attributes, and sensor-level event annotations curated from social-media traffic reports through a two-stage LLM-assisted pipeline. 
    \item We benchmark SpecLST on four real-world PeMS datasets against classical, graph-based, spectral, and Transformer-based baselines. The results show that SpecLST maintains competitive forecasting accuracy while reducing computational cost, especially in the multimodal PEMS-BAY22 setting. 
\end{itemize}

The remainder of this paper is organized as follows. Section~\ref{sec:related-work} reviews related work on graph-based, Transformer-based, and frequency-domain traffic forecasting methods. Section~\ref{sec:methodology} details the proposed SpecLST framework and its components. Section~\ref{sec:experiments} presents the experimental setup, results, parameter sensitivity analyses, and ablation studies. Section~\ref{sec:conclusion-future-work} concludes the paper and outlines directions for future research.

\section{Related Work}
\label{sec:related-work}

Traffic forecasting is a fundamental task in intelligent transportation systems. Existing approaches can be broadly categorized into traditional statistical methods and data-driven deep learning models \cite{traffic_deep_survey_2022,jiang2021gnn_survey}. Among the latter, spatio-temporal graph neural networks have become the dominant paradigm, while Transformer-based architectures have recently achieved notable gains in accuracy and modeling flexibility. 

\subsection{Graph-based spatio-temporal models}

Traffic forecasting over sensor networks is commonly formulated on a graph, where nodes correspond to detectors and edges encode physical proximity or learned dependencies. A major stream of research builds spatio-temporal neural networks by combining graph message passing with temporal sequence modeling.
DCRNN integrates diffusion convolution (random-walk-based graph filtering) into an RNN encoder-decoder to capture directed spatial propagation and temporal evolution \cite{li2018dcrnn}.
STGCN replaces recurrence with temporal gated convolutions and spatial graph convolutions, providing an efficient fully-convolutional alternative \cite{yu2018stgcn}.
ASTGCN introduces spatial and temporal attention to adaptively emphasize correlated sensors and time steps, and explicitly models multi-period traffic patterns \cite{guo2019astgcn}.

To reduce reliance on a predefined adjacency matrix, several studies have explored learning adaptive adjacency structures. Graph WaveNet learns an adaptive adjacency matrix via node embeddings and couples it with dilated temporal convolutions to capture long-range dependencies \cite{wu2019graphwavenet}.
AGCRN further learns node-adaptive parameters and constructs a data-adaptive graph, improving robustness to imperfect topology priors \cite{bai2020agcrn}.

However, diffusion-based graph models mainly describe traffic dynamics as propagation over the road network. This view can underrepresent node-intrinsic temporal patterns, such as local demand cycles or sensor-specific variations, that are not fully explained by neighboring sensors. To address this limitation, D2STGNN introduces a decoupled spatial-temporal framework with dynamic graph learning, achieving strong performance on multiple public benchmarks \cite{shao2022d2stgnn}. 

Overall, graph-based models have substantially advanced short-horizon accuracy, but most implementations still operate on limited temporal contexts and do not model heterogeneous exogenous signals such as weather, calendar, and incidents. 

\subsection{Transformer-based traffic forecasting}

Inspired by the success of self-attention in sequence modeling, Transformer-based architectures have recently emerged as a powerful alternative to convolutional and recurrent paradigms in traffic forecasting. By leveraging attention mechanisms, these models provide enhanced efficiency in capturing complex temporal dependencies and spatial interactions.

STTN \cite{sttn} integrates spatial and temporal attention within a unified Transformer framework, enabling joint modeling of node interactions and temporal dynamics. STGormer \cite{stgormer} further improves structural awareness by incorporating graph positional encoding into the attention computation, strengthening spatial dependency modeling.

To enhance adaptability to heterogeneous urban patterns, STAEformer \cite{staeformer} employs adaptive embeddings to dynamically adjust spatial representations. LSTTN \cite{lsttn} extends temporal receptive fields by incorporating long-horizon context modeling, demonstrating strong performance on datasets with extended historical inputs. More recently, PDFormer \cite{pdformer} introduces pattern-aware decomposition mechanisms to better capture diverse traffic regimes.

Overall, Transformer-based approaches have significantly improved predictive accuracy and modeling flexibility. However, most existing methods rely on full self-attention over raw temporal sequences, which lead to substantial computational overhead when handling long-term or multimodal inputs~\cite{fedformer}.

\subsection{Frequency-domain modeling for time series and traffic}

Frequency-domain modeling provides a complementary mechanism for representing periodicities and separating stable trends from high-frequency fluctuations. FEDformer incorporates seasonal-trend decomposition and Fourier-based frequency operators into a Transformer framework to better capture global structures and long-range dependencies with linear computational complexity \cite{fedformer}. Autoformer introduces an Auto-Correlation mechanism, efficiently computed via FFT, to capture period-based dependencies at the sub-series level while incorporating progressive series decomposition to improve long-term forecasting performance \cite{autoformer}.

In traffic forecasting, spectral decompositions are often used to disentangle trend and event-like variations. STWave leverages the discrete wavelet transform to decompose traffic time series into trend (low-frequency) and event (high-frequency) components, effectively converting temporal signals into a multi-resolution frequency representation. Each component is then processed through dual-channel spatio-temporal encoders, and spatial correlations are captured using Efficient Spectral Graph Attention Networks \cite{STWave}. By combining wavelet-based frequency decomposition with graph-based spatial modeling, STWave provides a compact and interpretable representation of both temporal periodicity and short-term fluctuations. Recent work has also attempted to unify spatial, temporal, and spectral modeling within a single attention framework. For example, STFAN introduces a space-time-frequency attention network that explicitly incorporates frequency-domain analysis into spatio-temporal learning, enabling joint modeling of hidden correlations across spatial, temporal, and spectral dimensions for improved mid- and long-term traffic forecasting~\cite{STFAN}.

However, existing frequency-domain traffic forecasting models primarily apply spectral transforms to traffic-only sequences over relatively short temporal windows. They rarely represent multi-week traffic history as a compact frequency-domain modality that can be jointly fused with heterogeneous exogenous information, such as weather, calendar, and incident signals. Consequently, the interaction between stable, long-term traffic patterns and short-term root-cause factors remains underexplored. 

In comparison with existing approaches, several important differences distinguish SpecLST from prior frequency-domain traffic models. STWave decomposes recent traffic into trend and event channels via wavelets, but its high-frequency component is processed by a parallel encoder of comparable depth to the main branch rather than a lightweight residual refinement mechanism, and its design is not centered on a frequency-domain long-history representation. FEDformer replaces self-attention with Fourier operators on relatively short input sequences, while Autoformer employs FFT-based auto-correlation within short temporal windows. Neither framework maps multi-week traffic history into a compact set of low-frequency spectral tokens for multimodal fusion. To the best of our knowledge, SpecLST is the first traffic forecasting framework that jointly integrates (i) frequency-domain long-history representation for multi-week contextual modeling, (ii) attention-based multimodal fusion with weather, calendar, and incident information, and (iii) a per-horizon high-frequency residual head within a unified architecture.

\section{Methodology}
\label{sec:methodology}

\subsection{Problem definition and preliminary}
We first define the notions of traffic network and multimodal data, and then formalize the traffic forecasting problem.

\paragraph{Definition 1. Traffic network} 

A traffic network is modeled as an undirected graph
$G = (V, E)$,
where $V$ denotes the set of nodes with cardinality $|V| = N$, and each node corresponds to a deployed sensor.
The edge set $E$ has cardinality $|E| = M$ and encodes the spatial connectivity between sensors. The strength of this connectivity is quantified by an adjacency matrix \mbox{$A \in \mathbb{R}^{N \times N}$}, whose entries give the pairwise connection weights between nodes.
The spatial position of each node is specified by its geographic coordinates 
\mbox{$\text{Loc} = (\text{lat}, \text{lon})$}.

\paragraph{Definition 2. Long-short term data}

In this study, we utilize long-term traffic features $X_{\text{Long}}$ and short-term multimodal data $X_{\text{Short}}$ as inputs to the proposed multimodal forecasting model, where long-term historical context helps capture underlying periodic traffic patterns and short-term data enhances the modeling of dynamic characteristics.

\begin{equation}
X_{\text{Long}}
= \big[ X_{t-L}, X_{t-L+1}, \ldots, X_{t-1} \big]
\end{equation}

\noindent where $X_{\text{Long}} \in \mathbb{R}^{L \times N}$ denotes the long-term historical traffic feature input, and $L$ represents the length of the long-term sequence.

\begin{equation}
X_{\text{Short}} = \big\{\, [X^{(d)}_{t-T}, X^{(d)}_{t-T+1}, \ldots, X^{(d)}_{t-1}] \ \big|\ d \in \mathcal{D} \,\big\},
\end{equation}

\noindent where the modality set $\mathcal{D}$ contains traffic, weather, and calendar, with events added when event data is available. $X_{\text{Short}}$ denotes the short-term multimodal features input, and $T$ represents the length of the short-term sequence. $X^{\text{traffic}} \in \mathbb{R}^{T \times N}$ represents the traffic features input.
$X^{\text{weather}} \in \mathbb{R}^{T \times W}$, where $W$ denotes the number of weather features.
$X^{\text{calendar}} \in \mathbb{R}^{T \times C}$, where $C$ denotes the number of calendar features.
$X^{\text{event}} \in \mathbb{R}^{T \times N}$, representing event features, is available only for datasets that provide sensor coordinates suitable for spatial alignment with incident reports.

Weather, calendar, and event inputs are treated as short-term contextual data because their effects are tied to the current forecasting horizon rather than to stable multi-week periodicity. Weather conditions can immediately alter road capacity and driving behavior, calendar attributes distinguish recurrent temporal regimes such as weekdays, weekends, and public holidays, and traffic events capture abrupt disruptions caused by incidents or road works. Combining these variables with recent traffic observations allows the model to explain short-term deviations from the long-term periodic baseline.

\paragraph{Definition 3. Traffic Forecasting} 

The traffic forecasting problem can be formulated as predicting traffic features, such as traffic flow or speed, $H$ time steps into the future. 
This prediction is based on a historical long-short term sequence and the adjacency matrix $A$:

\begin{equation}
X_{\text{Horizon}}
= f\!\left( A, X_{\text{Long}}, X_{\text{Short}} \right)
\end{equation}

\noindent where $X_{\text{Horizon}} \in \mathbb{R}^{N \times H}$ denotes the predicted traffic features. For example, when the data are sampled every 5 minutes and $H=12$, $X_{\text{Horizon}}$ contains the next 60 minutes of predicted flow or speed for all $N$ sensors.

\subsection{SpecLST Framework}

\begin{figure*}[t]
    \centering
    \includegraphics[width=0.8\textwidth]{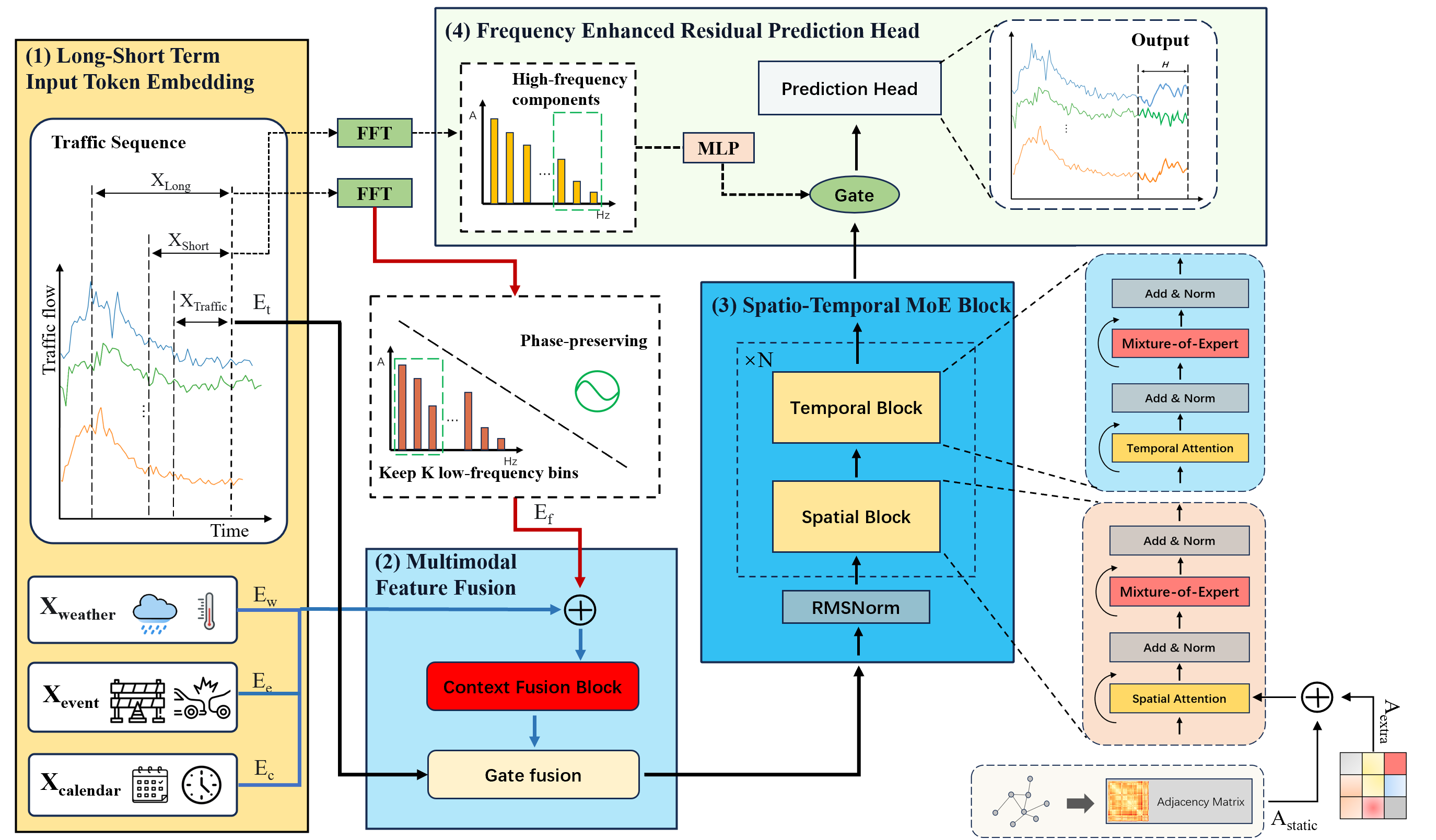}
    \caption{An overview of the SpecLST framework.}\label{framework}
\end{figure*}

The overall framework of SpecLST is illustrated in Fig.~\ref{framework}. The proposed SpecLST comprises four key components: (1) \emph{long-short term input token embedding}, (2) \emph{multimodal feature fusion}, (3) \emph{spatio-temporal MoE block}, and (4) \emph{frequency-enhanced residual prediction head}.

Specifically, long historical traffic is mapped by FFT into low-frequency tokens and aligned with the short-term window via phase rotation. Short-term modalities (traffic, weather, calendar, and optionally event) are embedded and fused with the FFT modality through attention-based context fusion. The fused embedding is passed to a stack of spatio-temporal blocks with alternating FFN and sparse MoE layers, where the spatial attention operates on the fused static and learnable adjacency. A frequency-enhanced residual head (FRH), driven by a separate recent window, produces a per-horizon high-frequency correction that is added to the backbone's prediction.

\subsubsection{Long-Short Term Input Token Embedding}
\label{sec:input_embedding}

The primary objective of the input embedding layer is to transform the heterogeneous and multimodal input data streams into a unified, high-dimensional state space. We categorize the input into three groups: (1) historical traffic features from a basic window of length $T$, (2) external context such as weather, calendar, and events, and (3) a low-frequency domain modality extracted from historical traffic features within a long window of length $L$.

Within the short-term feature set \( X_{\text{Short}} \), traffic data \( \mathbf{X}_{\text{traffic}} \), weather data \( \mathbf{X}_{\text{weather}} \), calendar data \( \mathbf{X}_{\text{calendar}} \), and event data \( \mathbf{X}_{\text{event}} \) are each projected into a high-dimensional state space through their respective linear embedding layers:

\begin{align}
\mathbf{E}_{\text{t}} &= \mathrm{Linear}_{\mathrm{t}}(\mathbf{X}_{\text{traffic}}), \\
\mathbf{E}_{\text{w}} &= \mathrm{Linear}_{\mathrm{w}}(\mathbf{X}_{\text{weather}}), \\
\mathbf{E}_{\text{c}} &= \mathrm{Linear}_{\mathrm{c}}(\mathbf{X}_{\text{calendar}}), \\
\mathbf{E}_{\text{e}} &= \mathrm{Linear}_{\mathrm{e}}(\mathbf{X}_{\text{event}}).
\end{align}

\noindent where $\mathbf{E}_{\text{t}}, \mathbf{E}_{\text{e}} \in \mathbb{R}^{B \times T \times N \times D}$, $\mathbf{E}_{\text{w}}, \mathbf{E}_{\text{c}} \in \mathbb{R}^{B \times T \times D}$, \( B \) is the batch size, \( D \) is the hidden dimension, and \( N \) is the number of sensors.

Observing long-period traffic is beneficial for improving a model's ability to capture stable traffic patterns. However, directly feeding such long raw sequences (e.g., more than a week of data) into attention-based architectures substantially increases parameter and memory requirements. To mitigate this issue, we define the long-term representation using the discrete Fourier transform (DFT) and compute its coefficients efficiently with FFT, retaining the first \(K\) low-frequency components. These low-frequency components capture slowly varying periodic patterns, such as daily and weekly traffic cycles, while requiring far fewer tokens than the original long sequence. Following the standard DSP convention, we use lowercase $x$ for time-domain signals and uppercase $X$ for their frequency-domain coefficients. The retained coefficients are defined as: 

\begin{align}
X^{(L)}_{b,k,n}
=\sum_{l=0}^{L_{\text{eff}}-1}{x}^{(L)}_{b,n}[l]\;
e^{-\mathrm{j}\,2\pi k l / L_{\text{eff}}} 
\end{align}

\noindent where $x^{(L)}_{b,n}[l]$ denotes the downsampled long-term traffic value of sample $b$ and sensor $n$ at index $l$. The stride $d_L$ is the temporal downsampling factor applied to the long-term sequence before FFT, yielding an effective sequence length $L_{\text{eff}} = L/d_L$. In our experiments we set $d_L = 2$, giving $L_{\text{eff}} = 2016$ for $L = 4032$ (two weeks of 5-minute samples). Here $b \in \{0,\ldots, B-1\}$, $n \in \{0,\ldots, N-1\}$, $k \in \{1,\ldots, K\}$, and $l \in \{0,\ldots, L_{\text{eff}} - 1\}$. The number of retained frequency bins $K$ is fixed for each dataset. The DC component ($k=0$) is intentionally excluded because it encodes the sensor-wise mean speed over the long window and carries no information about periodic structure. This constant offset is implicitly recovered through the residual connections in the backbone.

Two complementary operations are then applied to the retained spectrum. First, a frequency-wise amplitude normalization rescales each harmonic by the cross-sensor mean of that same harmonic. This removes raw scale differences between sensors (e.g., highway mainlines and on-ramps consistently exhibit different absolute energy at the daily and weekly harmonics) while preserving the \emph{relative} spectral shape that encodes periodic patterns. Second, each retained harmonic is projected back to the recent $T$ time steps through a phase rotation, which evaluates the inverse-DFT contribution of harmonic $k$ at temporal positions adjacent to the forecasting point. Together, these two operations turn $K$ complex coefficients into a $T$-length, $2K$-channel, phase-aware periodic embedding that aligns with and can be concatenated to the weather, calendar, event, and traffic embeddings.

We apply a frequency-wise amplitude normalization to normalize spectral energy across nodes:
\begin{align}
\widetilde{A}^{(L)}_{b,k,n}
&= \frac{A^{(L)}_{b,k,n}}
{\frac{1}{N}\sum_{n'=0}^{N-1} A^{(L)}_{b,k,n'} + \varepsilon}, \\
\widetilde{X}^{(L)}_{b,k,n}
&=\widetilde{A}^{(L)}_{b,k,n}\,
e^{\mathrm{j}\,\phi^{(L)}_{b,k,n}},
\end{align}

\noindent where $A^{(L)}_{b,k,n}=|X^{(L)}_{b,k,n}|$ denotes the magnitude, and $\phi^{(L)}_{b,k,n}$ is the intrinsic spectral phase obtained from the complex argument of the FFT output, and $\varepsilon$ is a small positive constant added for numerical stability.

Concretely, each retained harmonic is projected onto the recent $T$ time steps via the phase rotation defined below:
\begin{equation}
Z^{(L)}_{b,k,n}[t]
=
\widetilde{X}^{(L)}_{b,k,n}\,
e^{\mathrm{j}\,\tfrac{2\pi k (L_{\text{eff}}-T+t)}{L_{\text{eff}}}},
\qquad
t \in \{0,\ldots,T-1\}.
\end{equation}
\noindent where $t=0$ denotes the earliest step in the recent short-term window and $t=T-1$ denotes the step closest to the forecasting point.

The real and imaginary parts of $Z^{(L)}$ are concatenated along the frequency dimension to form the FFT features over the long-term window:
\begin{equation}
\mathbf{F}_{\text{fft}}=\operatorname{Concat}\!\big(\Re(Z^{(L)}),\, \Im(Z^{(L)})\big),
\end{equation}
where $\mathbf{F}_{\text{fft}} \in \mathbb{R}^{B\times T\times N\times 2K}$. This $(\text{Re}, \text{Im})$ representation is mathematically equivalent to retaining $(A_k\cos\phi_k, A_k\sin\phi_k)$ for each harmonic, but it allows the subsequent linear layer to learn arbitrary linear combinations of the two parts rather than committing to a particular trigonometric basis.

The resulting frequency spectral representation is used as an external modality, denoted as:
\begin{equation}
\mathbf{E}_{\text{f}} = \mathrm{Linear}\!\left(\mathbf{F}_{\text{fft}}\right),
\label{eq:E-f}
\end{equation}
with $\mathbf{E}_{\text{f}} \in \mathbb{R}^{B\times T\times N\times D}$.

\subsubsection{Multimodal Feature Fusion}

\label{sec:feature_fusion}

External temporal and low-frequency embeddings are fused via the multi-head self-attention mechanism and projected to the model's hidden dimension \( D \) by a context fusion module, forming the contextual features. Since $\mathbf{E}_{\text{w}}$ and $\mathbf{E}_{\text{c}}$ are global (sensor-agnostic) embeddings of shape $\mathbb{R}^{B \times T \times D}$, we broadcast them across the node dimension to obtain $\overline{\mathbf{E}}_{\text{w}}, \overline{\mathbf{E}}_{\text{c}} \in \mathbb{R}^{B \times T \times N \times D}$, so that all four modalities share the same shape. The four embeddings are then concatenated along the channel dimension and processed by multi-head self-attention:

\begin{equation}
\mathbf{E}_{\text{context}} = \mathrm{MHA}\!\left(\mathrm{Concat}_{\mathrm{ch}}\!\left[\overline{\mathbf{E}}_{\text{w}}, \overline{\mathbf{E}}_{\text{c}}, \mathbf{E}_{\text{e}}, \mathbf{E}_{\text{f}}\right]\right),
\end{equation}

\noindent where $\mathrm{Concat}_{\mathrm{ch}}$ denotes concatenation along the channel dimension, $\mathrm{MHA}(\cdot)$ denotes multi-head self-attention, and $\mathbf{E}_{\text{context}} \in \mathbb{R}^{B \times T \times N \times D}$.

Then, traffic embeddings are fused with contextual features using a cross-attention mechanism. The traffic embeddings serve as the query, while the context vectors form the key and value:

\begin{equation}
\mathbf{E}_{\text{aware}} = \mathrm{MHA}\!\left(\mathbf{E}_{\text{t}},\, \mathbf{E}_{\text{context}},\, \mathbf{E}_{\text{context}}\right),
\end{equation}

\noindent where the three arguments correspond to the query, key, and value, respectively.

A gating mechanism is then employed after the cross-attention step to balance the contributions of the traffic and context-aware embeddings:

\begin{align}
\mathbf{G} &= \sigma\!\left(\mathrm{Linear}\!\left(\mathrm{Concat}[\mathbf{E}_{\text{t}}, \mathbf{E}_{\text{aware}}]\right)\right), \\
\mathbf{E}_{\text{fused}} &= (1 - \mathbf{G}) \odot \mathbf{E}_{\text{t}} + \mathbf{G} \odot \mathbf{E}_{\text{aware}},
\end{align}
where \( \sigma \) is the sigmoid function, \( \odot \) denotes element-wise multiplication, and $\mathbf{G}, \mathbf{E}_{\text{fused}} \in \mathbb{R}^{B \times T \times N \times D}$. The gate $\mathbf{G}$ controls the proportion of the context-aware branch in the fused embedding: $\mathbf{G}\!\to\!0$ recovers the pure traffic embedding $\mathbf{E}_{\text{t}}$, whereas $\mathbf{G}\!\to\!1$ delegates the representation to the context-aware branch $\mathbf{E}_{\text{aware}}$. The bias of the gating $\mathrm{Linear}$ layer is initialized to zero so that $\mathbf{G}\approx 0.5$ at the start of training, ensuring that neither modality dominates before the model has learned the appropriate trade-off.

\subsubsection{Spatio-Temporal MoE Block}
\label{sec:st_moe}

Our backbone follows the general spatio-temporal coupling structure proposed in our prior research \cite{10.1007/978-981-95-4575-9_4} to model complex spatio-temporal dependencies. Each block adopts normalization and residual connections to stabilize deep representation learning. The temporal attention layer also uses multi-head self-attention with a causal lower-triangular mask, ensuring that the model's predictions at a given timestep depend only on itself and previous data, preserving autoregressive properties.

Relative to our prior STMMoE backbone \cite{10.1007/978-981-95-4575-9_4}, the components newly introduced in SpecLST are: (i) the FFT-based long-history branch and its phase-rotation alignment with the short-term window (Section~\ref{sec:input_embedding}), (ii) the learnable sparse graph augmentation $A_{\mathrm{extra}}$ on top of the static topology (this subsection), and (iii) the frequency-enhanced residual head with per-horizon gating (Section~\ref{sec:frh}). The multimodal fusion module and the alternating FFN/MoE arrangement within stacked spatio-temporal blocks follow our previous design.

For the spatial attention layer, we extend the spatial modeling from a static topology to a learnable one.
It dynamically constructs a spatial mask by fusing the static road network with a learned adjacency. Specifically, we let $A_{\mathrm{static}}\in\mathbb{R}^{N\times N}$ denote the static road-network adjacency. For all four datasets, $A_{\mathrm{static}}$ is constructed from pairwise sensor distances using a thresholded Gaussian kernel, following common practice in DCRNN-style preprocessing.
We introduce a global learnable sparse augmentation $A_{\mathrm{extra}} \in \mathbb{R}^{N \times N}$, parameterized as the outer product of two learnable node embedding vectors, that complements the road graph with a small number of additional connections.
To preserve structural sparsity and avoid dense attention, the augmentation is constrained to static non-edges and sparsified by retaining only the top-$k$ candidate edges per node, with $k=8$ in all experiments.
The "non-edge" constraint is a design choice rather than a strict structural requirement: it keeps the physical connectivity prior intact and assigns the learnable component the dedicated role of discovering non-local correlations (e.g., parallel corridors or shared upstream feeders) that the static topology does not encode. Down-weighting misleading static edges, where necessary, is handled implicitly by the attention layer that operates on top of $A_{\mathrm{fused}}$.
The fused adjacency is
\begin{equation}
A_{\mathrm{fused}} \;=\; A_{\mathrm{static}} \;+\; g \cdot A_{\mathrm{extra}},
\end{equation}
where $g = \sigma(\tilde{g}) \in (0,1)$ is a scalar learnable gate, parameterized by applying the sigmoid function to an unconstrained scalar parameter $\tilde{g}$. The parameter $\tilde{g}$ is initialized as $\operatorname{logit}(0.1)$ so that $g$ starts at $0.1$, making the spatial mask rely primarily on the static topology at the beginning of training and preventing the learned augmentation from dominating before it is sufficiently optimized.

The core of the temporal and spatial MoE module is the MoE layer, which enhances model capacity and efficiency. We adopt a sparse MoE rather than a dense feed-forward network of comparable parameter count because traffic flow exhibits regime heterogeneity (e.g., free-flow vs.\ congested, weekday peaks vs.\ weekend valleys, incident-affected vs.\ regular sensors) that benefits from conditional computation: only a subset of experts is activated per token, allowing different experts to specialize in different spatio-temporal regimes while keeping the per-sample FLOPs bounded. To balance model capacity and avoid excessive sparsity, we alternate between standard Feed-Forward Network layers and sparse MoE layers within the stacked spatio-temporal blocks. When the MoE layer is used, a gating network dynamically routes each input token to a small subset of expert subnetworks, and the final output is obtained as a weighted aggregation of the selected expert outputs. To further ensure stable sparse routing, an auxiliary load-balancing loss is incorporated during training. This regularization term is defined based on the routing importance and the realized expert load after capacity-constrained dispatch, encouraging balanced expert usage and reducing the risk of expert collapse.

Formally, the total training objective is
\begin{equation}
\mathcal{L} = \mathcal{L}_{\mathrm{MAE}} + \lambda_{\mathrm{aux}} \, \mathcal{L}_{\mathrm{aux}},
\end{equation}
where $\mathcal{L}_{\mathrm{MAE}} = \tfrac{1}{NH}\sum_{n,h}|\hat{X}_{n,h} - X_{n,h}|$ is the mean absolute error over all sensors and prediction horizons, and $\mathcal{L}_{\mathrm{aux}} = E \sum_{e=1}^{E} f_e \, p_e$ is the Switch-style load-balancing term computed as the product of the realized expert load $f_e$ (the fraction of tokens routed to expert $e$) and the mean routing probability $p_e$ assigned to expert $e$ over a batch. In all experiments we use $E=12$ experts with top-$k$ routing ($k=2$), an expert capacity factor of $1.25$, and $\lambda_{\mathrm{aux}}=0.01$.

\subsubsection{Frequency Enhanced Residual Prediction Head}
\label{sec:frh}

Short-term traffic variations often appear as high-frequency activity in recent observations, yet expanding the backbone's temporal receptive field to capture such dynamics is inefficient. To provide a lightweight correction signal, we introduce a frequency-enhanced residual head (FRH) that focuses specifically on high-frequency variations in a dedicated recent traffic window $W \in \mathbb{R}^{T_W \times N}$. Note that $W$ is a separate input slice from the short-term multimodal window of length $T$. It covers only recent traffic for the residual correction. Unlike trend-event wavelet decompositions such as STWave, which process the high-frequency component through a parallel encoder of comparable depth to the main branch, FRH is a lightweight, per-horizon-gated residual head that adjusts the final prediction without altering the spatio-temporal backbone. The combination of (i) a power-spectrum input slice, (ii) a horizon-specific reliability gate, and (iii) a recent-window scope that is distinct from the multimodal short-term window is, to our knowledge, new among traffic-forecasting residual heads. An FFT is applied to this window to obtain its frequency-domain representation, after which we take the power spectrum and slice it to keep only the upper-frequency band:
\begin{equation}
\mathbf{F}_{\mathrm{hf}} = \bigl(\, |\mathrm{FFT}(W)|^{2}_{k} \,\bigr)_{k \in \mathcal{H}}
\end{equation}
where \(\mathcal{H}\) denotes the selected high-frequency band obtained by keeping a fixed percentage of the highest frequencies, and the right-hand side stacks the power-spectrum entries indexed by $k \in \mathcal{H}$ into a vector.

The use of the power spectrum $|\mathrm{FFT}(W)|^{2}$ here, rather than the complex coefficients used in the long-history branch, reflects the different role of the two modules. The long-history branch encodes \emph{periodic patterns} whose phase carries timing information that must be preserved for alignment with the recent window. FRH, by contrast, is designed to flag the \emph{energy} of recent high-frequency activity (e.g., abrupt slowdowns) regardless of phase, since the magnitude alone is sufficient to indicate that a transient event is present, while the precise timing of that event has already been captured by the short-term traffic embedding. The high-frequency band $\mathcal{H}$ is taken as the upper third of the one-sided FFT bins.

The selected spectrum is then projected to a residual representation, and a horizon-specific learnable scalar $\beta_h$ is used to generate the gate that regulates its contribution:
\begin{align}
\Delta_h &= \mathrm{MLP}_h\!\left(\mathbf{F}_{\mathrm{hf}}\right), \\
G_h &= \sigma(\beta_h).
\end{align}

\noindent where $\Delta_h$ denotes the residual correction for prediction horizon $h$, and $G_h$ denotes the corresponding reliability gate.

Finally, the high-frequency residual produced by the MLP is injected into the model's prediction through a learnable gate, yielding the refined output at each prediction horizon $h$:
\begin{equation}
\widehat{X}_{\text{Horizon},h} = X_{\text{Horizon},h} + G_h \odot \Delta_h, \quad h \in \{3,6,12\}.
\label{eq:frh_fusion}
\end{equation}

FRH computes a lightweight residual for each sample, node, and prediction horizon using only a recent short window of traffic data, thereby incurring low computational cost. The resulting residual is further modulated by a learnable gate, ensuring that adjustments are only applied when high-frequency evidence is reliable. This design allows FRH to refine the backbone's predictions without altering its spatio-temporal structure or increasing its representational burden.

\section{Experiments}
\label{sec:experiments}

In this section, we evaluate SpecLST on four large real-world traffic datasets. We begin with the experimental setup, including the datasets, the implementation environment, and the hyperparameter configurations. We then report comparisons with baselines, followed by parameter sensitivity analyses and ablation studies.

\subsection{Datasets}

\subsubsection{Traffic data}

We evaluate our model on four public real-world traffic datasets, all collected from the Caltrans Performance Measurement System (PeMS) with a 5-minute recording interval (\cite{doi:10.3141/1748-12}).

\begin{enumerate}
    \item \text{PEMS-BAY} (\cite{PEMS-BAY}). This dataset consists of traffic speeds captured by 325 sensors on highways in the Bay Area. The data spans 6 months, from January~1,~2017 to May~31,~2017, totaling 52,116 time slices.

    \item \text{PEMS-BAY22} (\cite{PEMS-BAY-22}). This dataset consists of traffic flow and speed data collected from 325 sensors on highways in the Bay Area, spanning January~1,~2022 to June~30,~2022.

    \item \text{PEMS04} (\cite{PEMS04/08}). This dataset comprises flow, speed, and occupancy from 307 sensors situated on 29 highways in District~04, California. The dataset covers the period from January~1,~2018 to February~28,~2018, totaling 16,992 time slices.

    \item \text{PEMS08} (\cite{PEMS04/08}). This dataset includes three traffic features: flow, speed, and occupancy. It contains data from 170 sensors located on 8 highways in District~08, California. The data spans from July~1,~2016 to August~31,~2016, totaling 17,833 time slices.
\end{enumerate}

To provide a fair comparison, the dataset division in this paper is kept consistent with prior work in the literature. For PEMS-BAY, the ratio of the training, validation, and test sets is $7\!:\!2\!:\!1$, and for PEMS04, PEMS08, and PEMS-BAY22, it is $6\!:\!2\!:\!2$.
Among these benchmarks, PEMS-BAY22 provides the most recent Bay Area traffic observations in our evaluation.

\subsubsection{Context data}

To enhance the predictive capability of our model, we incorporate multimodal context data alongside traffic measurements, including weather conditions, calendar attributes, and traffic event information. These variables are aligned with the short-term traffic window because they explain immediate deviations from regular traffic periodicity: rainfall or snow can reduce capacity within the current horizon, weekends and holidays alter demand patterns, and incidents can create sudden congestion waves near affected sensors. Weather data includes local temperature, precipitation, snowfall, and categorical weather condition codes obtained from meteorological stations within the study area. Calendar attributes capture temporal patterns through day-of-week indicators, weekend flags, and public holiday markers for each timestamp.

Event data is extracted from incident reports posted on the X platform (formerly Twitter) by three public accounts focusing on Bay Area traffic: @511SFBay\footnote{\url{https://x.com/511SFBay}}, @caltransD4\footnote{\url{https://x.com/caltransD4}}, and @KCBSRadio\footnote{\url{https://x.com/KCBSRadio}}. We strategically limit integration of event data to the PEMS-BAY22 dataset because manual event annotation is labor-intensive.

To transform unstructured social media text into structured event embeddings, we design a two-stage pipeline combining automated filtering with rule-based classification. First, we employ a locally deployed large language model (\texttt{gpt-oss:20b}) to eliminate non-traffic content and extract location descriptors from tweet text. These descriptors are geocoded through OpenStreetMap's Nominatim service\footnote{\url{https://nominatim.openstreetmap.org/}}, with coordinates validated against the study area's bounding box to ensure spatial accuracy. Second, a deterministic regular expression-based classifier maps incident descriptions to a five-level severity scale ($L_0 \rightarrow L_4$), ranging from negligible disruptions ($L_0$) to major blockages ($L_4$). Events classified as $L_0$ (noise or non-impactful incidents) are discarded, while valid incidents ($L_1 \rightarrow L_4$) are spatiotemporally aligned with traffic sensor data using a 1.0~km spatial radius and 5-minute temporal window. This conservative approach prioritizes precision over recall to minimize false incident-flow associations. 

\subsection{Baseline Models}

We compare our method with a comprehensive set of classical and state-of-the-art baselines for traffic forecasting, including statistical models (\text{VAR} \cite{VAR_baselines} and \text{SVR} \cite{SVR_baselines}), graph-based recurrent networks (\text{DCRNN} \cite{li2018dcrnn}, \text{MTGNN} \cite{wu2020mtgnn}, \text{D$^2$STGNN} \cite{shao2022d2stgnn}, \text{STWave} \cite{STWave}), and recent Transformer-based spatio-temporal architectures (\text{STGormer} \cite{stgormer}, \text{LSTTN} \cite{lsttn}, \text{STTN} \cite{sttn}, \text{STAEformer} \cite{staeformer}).

\subsection{Experimental Environment and Hyperparameter Settings}

Our model is implemented in PyTorch, and the final reported experiments and efficiency measurements are conducted on a Linux server equipped with a 12-core AMD Ryzen 9 CPU and an NVIDIA GeForce RTX 4080 SUPER GPU with 16GB of memory. NVIDIA Brev cloud GPU resources obtained through the NVIDIA Academic Grant Program were used to accelerate our exploration and to support rerunning multiple baseline models across datasets.\footnote{\url{https://brev.nvidia.com/}}\footnote{\url{https://www.nvidia.com/en-us/industries/higher-education-research/academic-grant-program/}} 

All datasets are sampled at a 5-minute interval. Following the long-range design in our implementation, the long-term traffic context for the FFT modality is set to two weeks, corresponding to 4032 time steps. The short-term multimodal input is set to 1 hour, which corresponds to 12 time steps. For the frequency-enhanced residual head, the recent traffic window is set to $T_W=24$ steps, corresponding to 2 hours at the 5-minute sampling interval. The prediction horizons are set to $H\in\{3,6,12\}$, corresponding to 15, 30, and 60 minutes ahead, respectively. 

We train the model using the AdamW optimizer with an initial learning rate of $2.5\times10^{-4}$ and weight decay $0.01$. The main architectural hyperparameters follow our implementation: hidden dimension $128$, number of attention heads $8$, and $6$ spatio-temporal MoE blocks. The MoE module uses 12 experts with top-$k=2$ routing. 

Regarding the frequency-domain feature engineering, we distinguish between long-term periodicities and short-term transients through adaptive spectral thresholding. The demarcation between low- and high-frequency components is governed by the cumulative energy distribution of the power spectral density (PSD). Specifically, we perform an offline PSD analysis on the training dataset only, and select the smallest $K$ that preserves 95\% of the average cumulative spectral energy. Once determined, this dataset-specific $K$ is kept unchanged for all validation and test experiments on the same dataset. Since each dataset is trained with an independent model configuration, the subsequent linear embedding layer operates on a fixed-dimensional spectral input for that dataset.

To complement these stable patterns, we introduce a dedicated branch to capture incident-triggered anomalies, which are typically manifested as high-frequency residuals. We extract these cues from the recent traffic window used by the frequency-enhanced residual head. By applying an FFT to this localized sequence and filtering with a high-pass filter that retains the upper one-third of the one-sided FFT bins, the model isolates rapid fluctuations from the underlying trend. The one-third cutoff is chosen to lie above the empirical knee of the cumulative power spectrum on the training data, where short-term variability begins to dominate over the daily and weekly periodicity already represented by the long-history branch. These high-frequency components are then projected as residual signals to the prediction head, enhancing the framework's sensitivity to sudden traffic bursts or accidents that traditional attention mechanisms might smooth over.

\subsection{Experimental Results}

In this section, we evaluate the proposed model from both accuracy and efficiency perspectives. While several baselines achieve strong results on specific datasets, our model consistently maintains competitive predictive performance with lower computational cost.

\subsubsection{Prediction Accuracy}

Table~\ref{tab:main_results} provides a comprehensive performance comparison of all models across four real-world datasets for prediction horizons of 15, 30, and 60 minutes. Overall, SpecLST performs competitively across all datasets and horizons.

It is worth noting that the comparison with LSTTN reveals a dataset-dependent trade-off. LSTTN achieves the best or near-best results on PEMS-BAY and PEMS04, confirming that subseries-based long-short temporal modeling remains a strong solution on several established benchmarks. SpecLST, however, obtains the best results on PEMS-BAY22 and leads most metrics on PEMS08, showing that frequency-domain long-context modeling provides an effective alternative when extended historical context and heterogeneous contextual inputs are considered together. This difference is also consistent with the model mechanisms: LSTTN preserves long-term information through learned subseries-level temporal representations, whereas SpecLST maps long historical traffic into a fixed set of low-frequency coefficients before fusion. As a result, SpecLST avoids maintaining long temporal feature maps in the main forecasting backbone, which helps explain its more favorable memory and training-time profile in the Pareto analysis on PEMS-BAY22, where it achieves the lowest average MAE with lower resource consumption than LSTTN.

Traditional statistical and machine learning methods, such as VAR and SVR, yield the least favorable results. This is primarily because these models rely on strong assumptions of linearity or stationarity, which fail to capture the highly non-linear and non-stationary nature of urban traffic flow. Deep learning baselines that incorporate spatial dependencies, such as DCRNN and MTGNN, demonstrate marked improvements. By modeling the traffic network as a graph, they effectively capture inter-sensor correlations. More recent Transformer-based architectures, such as STTN, STGormer, STAEformer, and LSTTN, further improve prediction accuracy by capturing spatio-temporal dependencies.

\begin{table*}[!t]
\centering
\scriptsize
\caption{Prediction performance of traffic forecasting models on four datasets. Best results are in \textbf{bold}. Second-best results are \underline{underlined}.}
\label{tab:main_results}
\setlength{\tabcolsep}{5pt}
\renewcommand{\arraystretch}{1.2}
\begin{tabular}{l l ccc ccc ccc}
\toprule
\textbf{Datasets} & \textbf{Methods} &
\multicolumn{3}{c}{\textbf{Horizon 3}} &
\multicolumn{3}{c}{\textbf{Horizon 6}} &
\multicolumn{3}{c}{\textbf{Horizon 12}} \\
\cmidrule(lr){3-5}\cmidrule(lr){6-8}\cmidrule(lr){9-11}
& & MAE & RMSE & MAPE & MAE & RMSE & MAPE & MAE & RMSE & MAPE \\
\midrule
\multirow{12}{*}{\textbf{PEMS-BAY}}
  & VAR        & 1.90 & 4.13 & 3.92\% & 2.47 & 5.57 & 5.36\% & 3.42 & 7.60 & 7.75\% \\
  & SVR        & 1.52 & 3.55 & 3.48\% & 1.85 & 4.41 & 4.43\% & 2.34 & 5.61 & 6.00\% \\
  & DCRNN      & 1.36 & 2.90 & 2.90\% & 1.70 & 3.90 & 3.90\% & 2.07 & 4.82 & 4.90\% \\
  & MTGNN      & 1.33 & 2.79 & 2.81\% & 1.65 & 3.74 & 3.75\% & 1.94 & 4.49 & 4.62\% \\
  & D$^2$STGNN & 1.24 & 2.60 & 2.58\% & 1.55 & 3.52 & 3.49\% & 1.85 & 4.30 & 4.37\% \\
  & LSTTN      & \textbf{1.06} & \textbf{2.18} & \textbf{2.11\%} & \textbf{1.26} & \textbf{2.76} & \textbf{2.63\%} & \textbf{1.48} & \textbf{3.41} & \textbf{3.24\%} \\
  & STTN       & 1.27 & 2.70 & 2.61\% & 1.69 & 3.83 & 3.70\% & 2.26 & 5.18 & 5.39\% \\
  & STGormer   & 1.36 & 2.92 & 2.90\% & 1.69 & 3.92 & 3.81\% & 1.96 & 4.58 & 4.56\% \\
  & STAEformer & 1.31 & 2.78 & 2.76\% & 1.62 & 3.68 & 3.62\% & 1.88 & 4.34 & 4.41\% \\
  & STWave     & 1.42 & 3.00 & 3.09\% & 1.80 & 4.03 & 4.20\% & 2.18 & 4.94 & 5.43\% \\
\cmidrule(lr){2-11}
  & SpecLST       & \underline{1.16} & \underline{2.48} & \underline{2.36\%} & \underline{1.40} & \underline{3.19} & \underline{3.01\%} & \underline{1.68} & \underline{3.93} & \underline{3.78\%} \\
\midrule
\multirow{12}{*}{\textbf{PEMS-BAY22}}
  & VAR        & 20.96 & 35.89 & 16.22\% & 26.53 & 44.68 & 18.04\% & 38.00 & 60.65 & 26.39\% \\
  & SVR        & 21.79 & 34.43 & 16.22\% & 26.11 & 40.44 & 19.29\% & 36.30 & 55.25 & 26.05\% \\
  & DCRNN      & 17.59 & 33.89 & 12.65\% & 19.74 & 39.05 & 14.41\% & 22.38 & 43.37 & 18.33\% \\
  & MTGNN      & 15.06 & 24.10 & 10.58\% & 16.59 & 26.77 & 12.24\% & 18.77 & 30.14 & 13.00\% \\
  & D$^2$STGNN & 16.81 & 24.14 & 10.96\% & 18.97 & 26.71 & 10.64\% & 21.57 & 30.45 & 14.30\% \\
  & LSTTN      & \underline{13.65} & \underline{23.71} & \underline{9.48\%} & \underline{15.36} & \underline{26.60} & 11.56\% & \underline{17.10} & \underline{29.72} & 13.20\% \\
  & STTN       & 17.49 & 27.33 & 12.91\% & 21.13 & 33.23 & 15.49\% & 27.53 & 42.65 & 19.62\% \\
  & STGormer   & 15.21 & 24.64 & 10.93\% & 16.78 & 27.28 & 13.32\% & 19.53 & 31.19 & 16.18\% \\
  & STAEformer & 14.24 & 24.39 & 9.70\% & 15.60 & 27.20 & \underline{10.54\%} & 17.71 & 30.86 & \underline{12.15\%} \\
  & STWave     & 14.38 & 24.19 & 9.86\% & 15.91 & 26.95 & 10.88\% & 17.81 & 30.15 & 12.43\% \\
\cmidrule(lr){2-11}
  & SpecLST       & \textbf{13.19} & \textbf{23.17} & \textbf{9.41\%} & \textbf{14.13} & \textbf{25.39} & \textbf{10.06\%} & \textbf{15.61} & \textbf{28.54} & \textbf{11.26\%} \\
\midrule
\multirow{12}{*}{\textbf{PEMS04}}
  & VAR        & 23.79 & 37.18 & 14.23\% & 28.77 & 44.31 & 17.27\% & 39.67 & 59.72 & 24.15\% \\
  & SVR        & 27.80 & 43.70 & 19.71\% & 32.91 & 51.05 & 23.00\% & 43.80 & 67.24 & 30.04\% \\
  & DCRNN      & 20.34 & 31.94 & 13.65\% & 21.85 & 34.43 & 14.13\% & 26.97 & 41.11 & 16.84\% \\
  & MTGNN      & 17.97 & \underline{27.97} & 12.75\% & 18.73 & \underline{28.92} & 13.31\% & 19.73 & \underline{30.23} & 13.81\% \\
  & D$^2$STGNN & 17.44 & 28.64 & 11.64\% & 18.28 & 30.10 & 12.10\% & 19.55 & 31.99 & 12.82\% \\
  & LSTTN      & \underline{17.06} & \textbf{27.74} & 11.68\% & \textbf{17.62} & \textbf{28.67} & \underline{11.68\%} & \textbf{18.46} & \textbf{30.00} & \textbf{12.17\%} \\
  & STTN       & 19.72 & 31.68 & 12.85\% & 21.70 & 34.59 & 13.87\% & 25.17 & 39.35 & 15.68\% \\
  & STGormer   & 17.73 & 29.10 & 11.84\% & 18.61 & 30.61 & 12.39\% & 20.00 & 32.69 & 13.54\% \\
  & STAEformer & 17.49 & 28.60 & \underline{11.61\%} & 18.24 & 29.95 & 12.00\% & 19.37 & 31.71 & 12.83\% \\
  & STWave     & 17.55 & 28.83 & 11.82\% & 18.42 & 30.25 & 12.63\% & 19.70 & 32.14 & 13.37\% \\
\cmidrule(lr){2-11}
  & SpecLST       & \textbf{16.94} & 28.17 & \textbf{11.07\%} & \underline{17.74} & 29.47 & \textbf{11.57\%} & \underline{18.92} & 31.37 & \underline{12.34\%} \\
\midrule
\multirow{12}{*}{\textbf{PEMS08}}
  & VAR        & 18.90 & 29.55 & 10.52\% & 22.94 & 35.79 & 12.74\% & 32.27 & 48.93 & 17.92\% \\
  & SVR        & 22.88 & 37.08 & 12.85\% & 26.94 & 42.62 & 15.13\% & 35.76 & 55.03 & 20.06\% \\
  & DCRNN      & 15.64 & 25.48 & 10.04\% & 17.69 & 27.63 & 11.38\% & 22.51 & 34.21 & 14.17\% \\
  & MTGNN      & 13.55 & \underline{21.05} & 9.99\% & 14.08 & \textbf{22.10} & 10.50\% & 14.96 & \underline{23.58} & 10.96\% \\
  & D$^2$STGNN & 13.14 & 21.42 & 8.55\% & 14.21 & 23.65 & 9.12\% & 15.73 & 26.41 & 10.17\% \\
  & LSTTN      & 13.17 & 21.89 & 8.63\% & 13.71 & 22.89 & 9.09\% & \underline{14.54} & \textbf{23.47} & \underline{9.77\%} \\
  & STTN       & 16.48 & 25.36 & 9.48\% & 19.42 & 29.58 & 10.78\% & 25.65 & 37.43 & 13.72\% \\
  & STGormer   & 13.03 & 22.43 & 8.51\% & 13.92 & 24.48 & 8.99\% & 15.21 & 26.95 & 9.86\% \\
  & STAEformer & 12.87 & 21.60 & \underline{8.33\%} & \underline{13.56} & 23.44 & \underline{8.90\%} & 14.91 & 25.91 & 9.86\% \\
  & STWave     & \underline{12.71} & 21.90 & 8.61\% & 13.57 & 23.84 & 8.97\% & 14.81 & 26.07 & 9.84\% \\
\cmidrule(lr){2-11}
  & SpecLST       & \textbf{12.39} & \textbf{20.84} & \textbf{7.96\%} & \textbf{13.16} & \underline{22.35} & \textbf{8.44\%} & \textbf{14.34} & 24.26 & \textbf{9.24\%} \\
\bottomrule
\end{tabular}
\end{table*}

\subsection{Sensitivity Analysis}

Table~\ref{tab:sensitivity_delta} summarizes the sensitivity results on PEMS-BAY22 as absolute differences from the corresponding reference settings.

\begin{table*}[!t]
\centering
\caption{Sensitivity results on PEMS-BAY22 reported as absolute differences from the corresponding reference settings. Negative values indicate improved performance, and $\Delta$MAPE is reported in percentage points. Best results within each group and column are in \textbf{bold}.}
\label{tab:sensitivity_delta}
\scriptsize
\setlength{\tabcolsep}{3.2pt}
\renewcommand{\arraystretch}{1.05}
\begin{tabular}{l c ccc ccc ccc}
\toprule
\multirow{2}{*}{\textbf{Reference}} & \multirow{2}{*}{\textbf{Setting}} &
\multicolumn{3}{c}{\textbf{H3}} &
\multicolumn{3}{c}{\textbf{H6}} &
\multicolumn{3}{c}{\textbf{H12}} \\
\cmidrule(lr){3-5}\cmidrule(lr){6-8}\cmidrule(lr){9-11}
& & $\Delta$MAE & $\Delta$RMSE & $\Delta$MAPE &
$\Delta$MAE & $\Delta$RMSE & $\Delta$MAPE &
$\Delta$MAE & $\Delta$RMSE & $\Delta$MAPE \\
\midrule
\multirow{3}{*}{$\Delta$ to $BS=4$}
& 8  & \textbf{$-0.054$} & \textbf{$-0.126$} & \textbf{$-0.23$} & \textbf{$-0.063$} & \textbf{$-0.130$} & \textbf{$-0.18$} & \textbf{$-0.090$} & \textbf{$-0.117$} & \textbf{$-0.35$} \\
& 16 & $+0.016$ & $-0.074$ & $-0.08$ & $+0.012$ & $-0.076$ & $+0.02$ & $+0.029$ & $-0.060$ & $-0.06$ \\
& 32 & $+0.063$ & $-0.045$ & $+0.04$ & $+0.044$ & $-0.066$ & $+0.04$ & $+0.056$ & $0.000$ & $-0.13$ \\
\midrule
\multirow{3}{*}{$\Delta$ to $D=64$}
& 128 & $-0.145$ & $-0.083$ & $-0.31$ & $-0.210$ & $-0.142$ & $-0.50$ & $-0.259$ & $-0.187$ & $-0.56$ \\
& 256 & $-0.154$ & $-0.096$ & \textbf{$-0.40$} & $-0.241$ & $-0.197$ & \textbf{$-0.56$} & $-0.307$ & $-0.235$ & \textbf{$-0.64$} \\
& 512 & \textbf{$-0.199$} & \textbf{$-0.161$} & $-0.27$ & \textbf{$-0.286$} & \textbf{$-0.272$} & $-0.41$ & \textbf{$-0.360$} & \textbf{$-0.403$} & $-0.46$ \\
\midrule
\multirow{6}{*}{$\Delta$ to $ST=1$}
& 2  & $-0.142$ & \textbf{$-0.198$} & $-0.03$ & $-0.129$ & $-0.166$ & $+0.01$ & $-0.197$ & $-0.260$ & $-0.09$ \\
& 4  & $-0.148$ & $-0.119$ & $-0.16$ & $-0.180$ & $-0.128$ & $-0.16$ & $-0.253$ & $-0.181$ & $-0.14$ \\
& 6  & \textbf{$-0.228$} & $-0.192$ & \textbf{$-0.31$} & \textbf{$-0.261$} & \textbf{$-0.212$} & \textbf{$-0.30$} & \textbf{$-0.364$} & \textbf{$-0.343$} & \textbf{$-0.37$} \\
& 8  & $-0.073$ & $+0.106$ & $-0.18$ & $-0.081$ & $+0.167$ & $-0.20$ & $-0.181$ & $+0.112$ & $-0.34$ \\
& 10 & $-0.151$ & $-0.058$ & $-0.16$ & $-0.105$ & $+0.038$ & $-0.06$ & $-0.173$ & $-0.008$ & $-0.03$ \\
& 12 & $-0.141$ & $-0.094$ & $-0.23$ & $-0.188$ & $-0.150$ & $-0.20$ & $-0.269$ & $-0.267$ & $-0.28$ \\
\bottomrule
\end{tabular}
\end{table*}

As shown in Table~\ref{tab:sensitivity_delta}, $BS=8$ yields consistent error reductions across all horizons and metrics relative to $BS=4$, whereas larger batch sizes produce mixed changes. Increasing the hidden dimension generally improves MAE and RMSE, with $D=512$ producing the largest reductions, while $D=256$ performs best in MAPE. Nevertheless, the additional gains beyond $D=128$ are metric-dependent, and $D=128$ is retained as a compact operating point to limit model size and computational cost. Regarding model depth, $ST=6$ provides the strongest overall improvement, whereas deeper configurations yield inconsistent gains and occasionally increase RMSE. Accordingly, we use $BS=8$, $D=128$, and $ST=6$ in the reported model configuration.

\subsection{Ablation Studies}

All ablation studies are conducted on the PEMS-BAY22 dataset, with results reported as averages over three random seeds. By systematically removing individual components or input groups, we evaluate their impact on predictive accuracy. The tested settings are defined as follows:

\begin{itemize}[leftmargin=1.5em]
  \item \textit{w/o LAM}: Removal of the learnable adjacency matrix, retaining only the static adjacency matrix derived from the road network.
  \item \textit{w/o AM}: Removal of the adjacency matrix.
  \item \textit{w/o CA}: Removal of calendar features from the exogenous inputs.
  \item \textit{w/o WE}: Removal of weather features from the exogenous inputs.
  \item \textit{w/o EV}: Removal of event features from the exogenous inputs.
  \item \textit{traffic only}: Removal of all exogenous modalities, leaving only traffic observations as short-term inputs.
  \item \textit{w/o FRH}: Removal of the frequency-enhanced residual head.
  \item \textit{w/o FFT}: Removal of the long-history FFT branch.
  \item \textit{w/o Spectral}: Removal of both spectral components, namely the FFT branch and frequency-enhanced residual head.
\end{itemize}

Table~\ref{tab:ablation_adj} examines the role of adjacency modeling. The learned adjacency yields the best results across all horizons. This indicates that the physical graph prior remains useful, but augmenting it with a sparse learnable structure better captures task-dependent non-local correlations that are not fully encoded by the original road network.

\begin{table*}[!t]
\centering
\caption{Adjacency matrix ablation experiment results. Best results per column are in \textbf{bold}.}
\label{tab:ablation_adj}
\setlength{\tabcolsep}{6pt}
\footnotesize
\begin{tabular}{lccccccccc}
\toprule
& \multicolumn{3}{c}{Horizon 3} & \multicolumn{3}{c}{Horizon 6} & \multicolumn{3}{c}{Horizon 12} \\
\cmidrule(lr){2-4}\cmidrule(lr){5-7}\cmidrule(lr){8-10}
Method & MAE & RMSE & MAPE & MAE & RMSE & MAPE & MAE & RMSE & MAPE \\
\midrule
w/o LAM & 13.42 & 23.41 & 9.66\% & 14.42 & 25.71 & 10.35\% & 16.01 & 28.96 & 11.64\% \\
w/o AM     & 13.46 & 23.51 & 9.57\% & 14.46 & 25.84 & 10.23\% & 16.04 & 29.06 & 11.52\% \\
Original       & \textbf{13.19} & \textbf{23.17} & \textbf{9.41\%} & \textbf{14.13} & \textbf{25.39} & \textbf{10.06\%} & \textbf{15.61} & \textbf{28.54} & \textbf{11.26\%} \\
\bottomrule
\end{tabular}
\end{table*}

Table~\ref{tab:ablation_exogenous} evaluates the contribution of exogenous modalities. Removing all contextual modalities produces the clearest degradation, confirming that multimodal information provides complementary cues beyond historical traffic alone. Among individual modalities, event information generally has the strongest influence on overall predictive performance, while weather and calendar features also contribute positively but with comparatively smaller effects. Moreover, the trends in RMSE and MAPE are not entirely identical to those in MAE, suggesting that different exogenous modalities affect different aspects of prediction quality rather than only improving average absolute error.

\begin{table*}[!t]
\centering
\caption{Exogenous modality ablation experiment results. Best results per column are in \textbf{bold}.}
\label{tab:ablation_exogenous}
\setlength{\tabcolsep}{6pt}
\footnotesize
\begin{tabular}{lccccccccc}
\toprule
& \multicolumn{3}{c}{Horizon 3} & \multicolumn{3}{c}{Horizon 6} & \multicolumn{3}{c}{Horizon 12} \\
\cmidrule(lr){2-4}\cmidrule(lr){5-7}\cmidrule(lr){8-10}
Method & MAE & RMSE & MAPE & MAE & RMSE & MAPE & MAE & RMSE & MAPE \\
\midrule
w/o CA    & 13.23 & 23.21 & 9.45\% & 14.19 & 25.39 & 10.13\% & 15.69 & 28.62 & 11.34\% \\
w/o WE     & 13.22 & 23.23 & 9.53\% & 14.18 & 25.47 & 10.18\% & 15.73 & 28.62 & 11.49\% \\
w/o EV       & 13.26 & 23.24 & 9.43\% & 14.23 & 25.52 & 10.12\% & 15.79 & 28.77 & 11.40\% \\
traffic only  & 13.35 & 23.27 & 9.51\% & 14.32 & 25.55 & 10.21\% & 15.88 & 28.77 & 11.46\% \\
Original      & \textbf{13.19} & \textbf{23.17} & \textbf{9.41\%} & \textbf{14.13} & \textbf{25.39} & \textbf{10.06\%} & \textbf{15.61} & \textbf{28.54} & \textbf{11.26\%} \\
\bottomrule
\end{tabular}
\end{table*}

Table~\ref{tab:ablation_spectral} isolates the contribution of different spectral components. The results show that removing any spectral component leads to performance degradation, demonstrating the benefit of spectral modeling for traffic prediction. Among the ablation settings, the model without FFT shows the greatest overall deterioration, highlighting the importance of frequency-domain information. In this ablation, w/o FFT means removing the long-history frequency branch and retaining the remaining short-term forecasting backbone, rather than replacing FFT with raw multi-week attention. This design choice is necessary because direct attention over raw long-history traffic has a memory cost growing quadratically with $L$. SpecLST instead computes the FFT in $\mathcal{O}(L\log L)$, keeps $K\ll L$ low-frequency coefficients, and applies attention only over the short window $T=12$, allowing multi-week historical context to be incorporated without raw long-sequence attention. Meanwhile, removing FRH or the spectral module also causes consistent, though comparatively milder, declines, suggesting that these components provide complementary improvements to the final performance.

\begin{table*}[!t]
\centering
\caption{Spectral component ablation experiment results. Best results per column are in \textbf{bold}.}
\label{tab:ablation_spectral}
\setlength{\tabcolsep}{6pt}
\footnotesize
\begin{tabular}{lccccccccc}
\toprule
& \multicolumn{3}{c}{Horizon 3} & \multicolumn{3}{c}{Horizon 6} & \multicolumn{3}{c}{Horizon 12} \\
\cmidrule(lr){2-4}\cmidrule(lr){5-7}\cmidrule(lr){8-10}
Method & MAE & RMSE & MAPE & MAE & RMSE & MAPE & MAE & RMSE & MAPE \\
\midrule
w/o FRH           & 13.23 & \textbf{23.16} & 9.43\% & 14.15 & 25.41 & \textbf{10.03\%} & 15.66 & 28.63 & 11.27\% \\
w/o FFT           & 13.48 & 23.47 & 9.61\% & 14.55 & 25.86 & 10.34\% & 16.33 & 29.34 & 11.71\% \\
w/o Spectral  & 13.61 & 23.60 & 9.77\% & 14.68 & 26.03 & 10.47\% & 16.55 & 29.67 & 11.94\% \\
Original              & \textbf{13.19} & 23.17 & \textbf{9.41\%} & \textbf{14.13} & \textbf{25.39} & 10.06\% & \textbf{15.61} & \textbf{28.54} & \textbf{11.26\%} \\
\bottomrule
\end{tabular}
\end{table*}

Overall, the ablation results suggest that the long-history FFT branch provides the largest observed contribution to prediction accuracy, followed by the exogenous context inputs, especially event information. The learnable adjacency and FRH provide smaller but consistent gains, indicating that they mainly refine spatial dependency modeling and short-term transient correction.

\subsection{Efficiency}

To evaluate the practical utility of SpecLST, we perform a Pareto analysis on the PEMS-BAY22 dataset, comparing the trade-off between forecasting accuracy and resource consumption. The MAE value in this analysis is averaged over the three forecasting horizons $H\in\{3,6,12\}$.

\begin{figure*}[!t]
    \centering
    \subfloat[Average MAE vs.\ memory usage.]{%
        \includegraphics[width=0.48\textwidth]{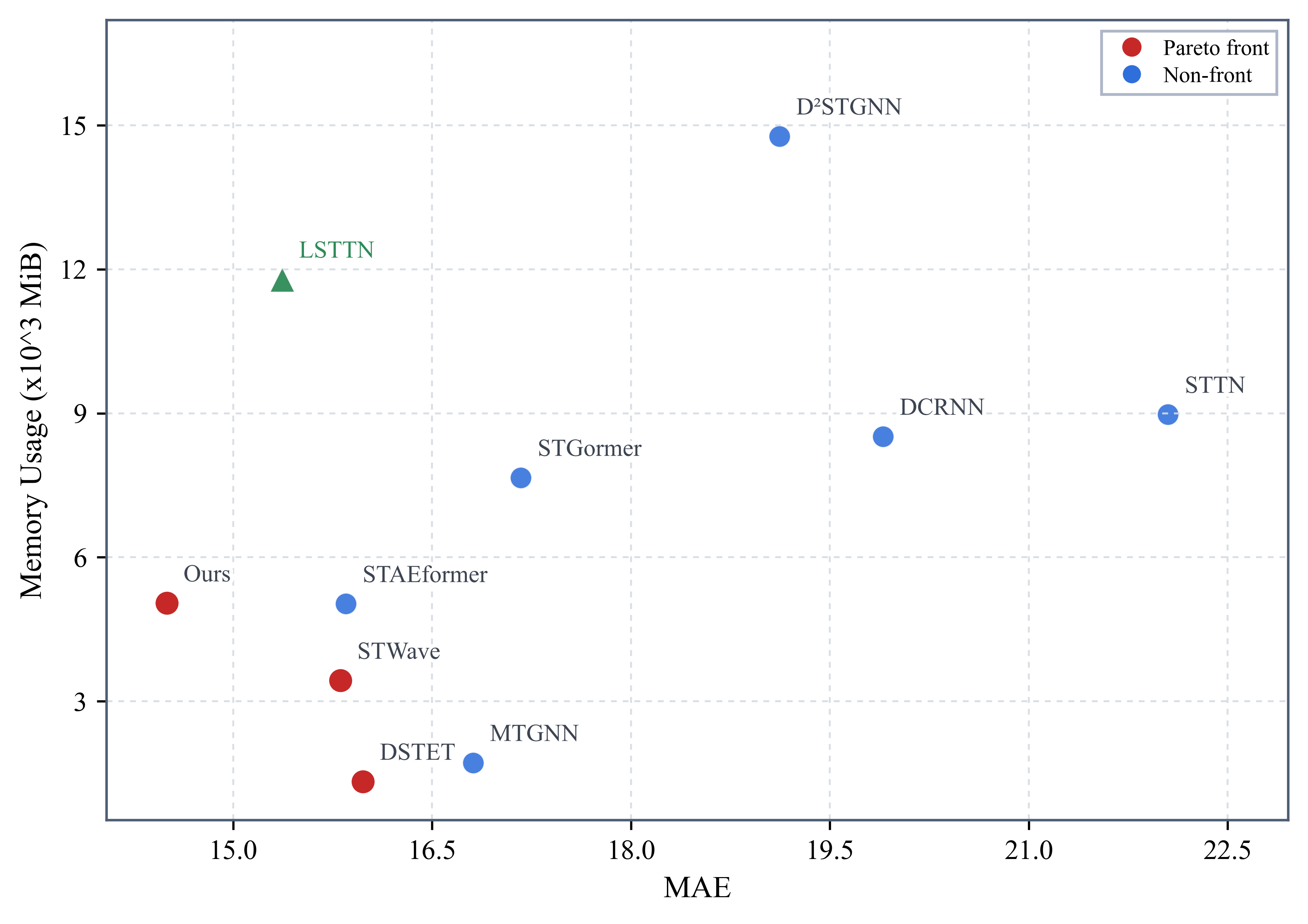}%
        \label{fig:pareto_memory}}
    \hfill
    \subfloat[Average MAE vs.\ training time.]{%
        \includegraphics[width=0.48\textwidth]{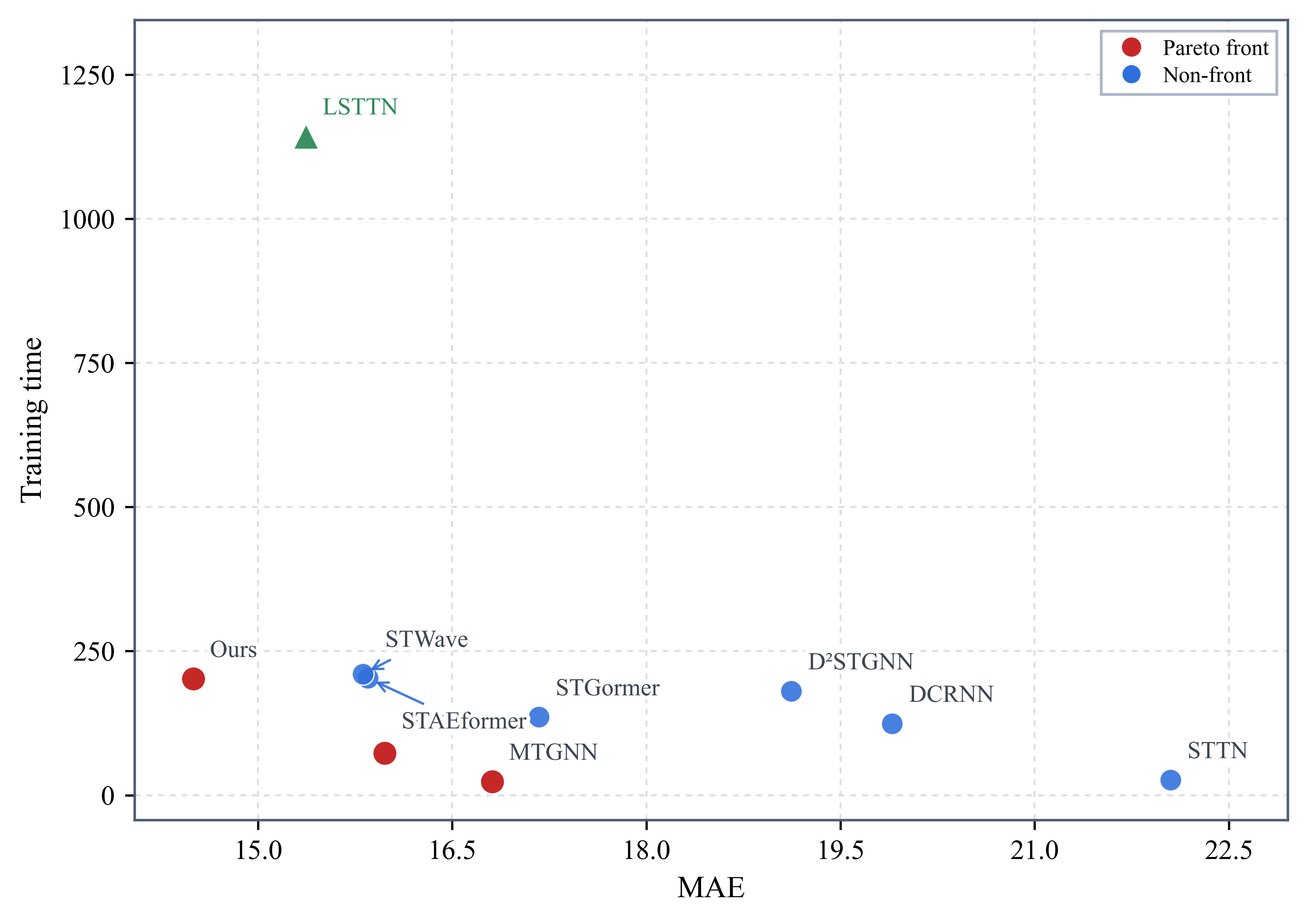}%
        \label{fig:pareto_time}}
    \caption{Efficiency comparison using Pareto frontier analysis on PEMS-BAY22.}\label{fig:pareto} 
\end{figure*}

In a Pareto analysis, lower-left points are preferable because they simultaneously reduce prediction error and resource usage. A model is on the Pareto frontier if no other model achieves both lower MAE and lower resource consumption. In Fig.~\ref{fig:pareto}, red circles denote Pareto-front models, blue circles denote non-front models, and LSTTN is marked separately with a green triangle. In the memory-usage panel of Fig.~\ref{fig:pareto}, SpecLST lies at the accuracy-oriented end of the frontier: it achieves the lowest average MAE while maintaining a moderate memory footprint, despite using multimodal inputs and two weeks of historical context. STWave and DSTET occupy lower-memory frontier positions, but with higher prediction error. By contrast, LSTTN is highlighted as a non-front model: although it remains a strong accuracy competitor, it requires substantially more memory than SpecLST because its long-context module still maintains temporal feature maps derived from long historical sequences. Thus, SpecLST provides a stronger overall trade-off when accuracy, multimodal context, and memory feasibility are considered together.

Regarding temporal efficiency, the training-time panel of Fig.~\ref{fig:pareto} shows a similar pattern. SpecLST remains the accuracy-oriented frontier point, while faster frontier models such as MTGNN achieve lower training time at the cost of higher prediction error. LSTTN achieves competitive accuracy but is far from the frontier because it requires a much longer training time due to its long-context temporal module. This indicates that the FFT representation does not merely reduce memory pressure. It also makes long-history multimodal forecasting practical without moving to the high-time regime of raw long-sequence models.

\section{Conclusion and Future Work}
\label{sec:conclusion-future-work}

In this paper, we propose SpecLST, a frequency-enhanced long-short term spatio-temporal multimodal mixture-of-experts framework designed to address the challenges of high-dimensional multimodal fusion and long-range historical dependency modeling in urban traffic forecasting. By representing long historical traffic in the frequency domain through FFT, the model effectively captures periodic signatures from several weeks of context with low memory usage, bypassing the computational bottlenecks inherent in traditional self-attention mechanisms over raw long sequences. Experimental results across four real-world benchmarks show that the combination of frequency-domain long-context representation, attention-based context fusion, and high-frequency residual correction yields competitive predictive performance and favorable resource efficiency. 


Empirically, SpecLST achieves the best results across all reported metrics and horizons on PEMS-BAY22, leads most metrics on PEMS08, and remains competitive with the strongest baselines on PEMS-BAY and PEMS04. It achieves the lowest average MAE while sitting on the favorable end of the Pareto frontier for both memory usage and training time, indicating that the frequency-domain long-history representation supports accuracy gains without the memory penalty of raw long-sequence attention. Ablation studies further confirm that each spectral and contextual component contributes a complementary signal, with the long-history FFT branch and the event modality providing the most substantial individual gains. 


Several directions remain open for future research. First, the learnable graph augmentation in SpecLST is global and time-invariant. Replacing the outer-product parameterization of $A_{\mathrm{extra}}$ with an attention-derived adjacency conditioned on the latest $T$ steps would allow the spatial structure to evolve in real time with traffic and capture transient correlations during peak hours or incidents. Second, beyond FFT, alternative long-history representation techniques, including wavelet transforms, discrete cosine transform (DCT)-based representations, or neural sequence summarization, may offer complementary ways to encode long-range context and non-stationary trends in heterogeneous modalities. Third, deep learning models struggle with cross-region generalization because their data-dependence inevitably reflects regional biases in network topology, mobility patterns, and data availability. Recent large language models have demonstrated strong generalization and zero-shot reasoning capabilities, which may provide useful priors for future traffic forecasting systems that need to adapt across regions with limited local supervision.

\section*{Acknowledgment}
The authors acknowledge NVIDIA for providing Brev cloud GPU resources through the NVIDIA Academic Grant Program, which supported SpecLST model exploration and baseline reruns.

%

\bibliographystyle{IEEEtran}
\bibliography{cas-refs}

\end{document}